\documentclass{article}
\usepackage{longtable,booktabs,array}
\usepackage{calc} 
\usepackage{etoolbox}
\makeatletter
\patchcmd\longtable{\par}{\if@noskipsec\mbox{}\fi\par}{}{}
\makeatother
\IfFileExists{footnotehyper.sty}{\usepackage{footnotehyper}}{\usepackage{footnote}}
\makesavenoteenv{longtable}\usepackage{longtable,booktabs,array}
\usepackage{calc} 
\usepackage{etoolbox}
\usepackage{caption}
\usepackage{subcaption}
\makeatletter
\patchcmd\longtable{\par}{\if@noskipsec\mbox{}\fi\par}{}{}
\makeatother
\IfFileExists{footnotehyper.sty}{\usepackage{footnotehyper}}{\usepackage{footnote}}
\makesavenoteenv{longtable}
\usepackage[english]{babel}
\usepackage{listings}
\usepackage{color}
\definecolor{dkgreen}{rgb}{0,0.6,0}
\definecolor{gray}{rgb}{0.5,0.5,0.5}
\definecolor{mauve}{rgb}{0.58,0,0.82}
\usepackage{amsmath}
\usepackage{graphicx}
\usepackage[colorlinks=true, allcolors=blue]{hyperref}
\usepackage{geometry}

\usepackage{lineno}

\title{A Timeline and Analysis  for Representation Plasticity in Large Language Models}
\author{Akshat Kannan}

\begin{document}
\maketitle

\begin{abstract}
The ability to steer AI behavior is crucial to preventing its long term dangerous and catastrophic potential. Representation Engineering (RepE) has emerged as a novel, powerful method to steer internal model behaviors, such as "honesty”, at a top-down level. Understanding the steering of representations should thus be placed at the forefront of alignment initiatives. Unfortunately, current efforts to understand plasticity at this level are highly neglected. This paper aims to bridge the knowledge gap and understand how LLM representation stability, specifically for the concept of “honesty”, and model plasticity evolve by applying steering vectors extracted at different fine-tuning stages, revealing differing magnitudes of shifts in model behavior. The findings are pivotal, showing that while early steering exhibits high plasticity, later stages have a surprisingly responsive critical window. This pattern is observed across different model architectures, signaling that there is a general pattern of model plasticity that can be used for effective intervention. These insights greatly contribute to the field of AI transparency, addressing a pressing lack of efficiency limiting our ability to effectively steer model behavior. Full code developed for the project can be found at
\url{https://github.com/UltraTsar/NonTrivialRepE\_Timeline/tree/main}.
\end{abstract}

\section{Overview of Current Literature}\label{h.m8x7orvex0ri}

Large language models (LLMs) have transformed the landscape of natural language processing (NLP). Despite their successes, a deeper understanding of how LLMs acquire, refine, and stabilize internal representations during fine-tuning remains underdeveloped in the field. The concept of plasticity—the model's proneness or ability to change behavior—has significant implications for both theory and practice in the field of machine learning. In neural networks, the degree of plasticity influences how well models can learn new tasks and model rigidity \cite{lyle_2023_understanding}.

Recent work in Representation Engineering (RepE) has sought to address this knowledge gap by offering a methodology to both probe and steer the internal representations of LLMs \cite{zou_2023_representation}. RepE involves the extraction of steering vectors, which are latent directions in the representation space that can be modified to influence model behavior. 

This research builds on these developments, hoping to analyze LLM plasticity over the course of fine-tuning. I specifically analyze how steering interventions vary in their effectiveness across different stages of fine-tuning.

The study is grounded in recent advancements in model interpretability and RepE \cite{liu_2023_incontext, cao_2024_personalized}. Notably, I draw inspiration from frameworks analyzing representational geometry, which enable us to trace how specific steering vectors affect semantic alignment and, more importantly, behavioral shifts in LLMs over time. 

\section{Overview of Representation Engineering}\label{h.3yx2i53hn6l5}

Zou et al's journal paper \cite{zou_2023_representation} provides a detailed overview of RepE strategy. It entails two main steps: probing (extraction) and steering. 

\subsection{Probing}\label{h.llgbxe31y8hw}

During probing, sets of prompts are used to extract “concept” vectors from the neural activity of the model. Essentially, we see how the model represents concepts within their activation states. 

The key step of RepE concept vector extraction is to subtract the complement of a concept from the concept (i.e. the concept vector of honesty would be the subtraction of the vector representative of dishonesty from honesty). 

The method of extraction in this paper will be to calculate the mean activation states of the last hidden layer given many prompts for both honest and dishonest scenarios. 

\subsection{Steering}\label{h.afmley53vjgv}

After acquiring the extracted representation as a concept/steering vector, to steer the model we must apply it. To do this, there are multiple methods. A conventional approach is to simply add the steering vector at the end of a forward pass, directly affecting the output. However, since we are seeking to understand how steering affects training dynamics as well, we employ a different approach.

After acquiring the steering vector, I recalculate loss using the steering vector through a process akin to regularization (detailed in \hyperref[h.69wq33xa2orv]{3.4}). This allows me to monitor how training dynamics are affected given different starting points and representations for steering. 

\begin{figure}[h]
    \centering
    \includegraphics[width=0.5\linewidth]{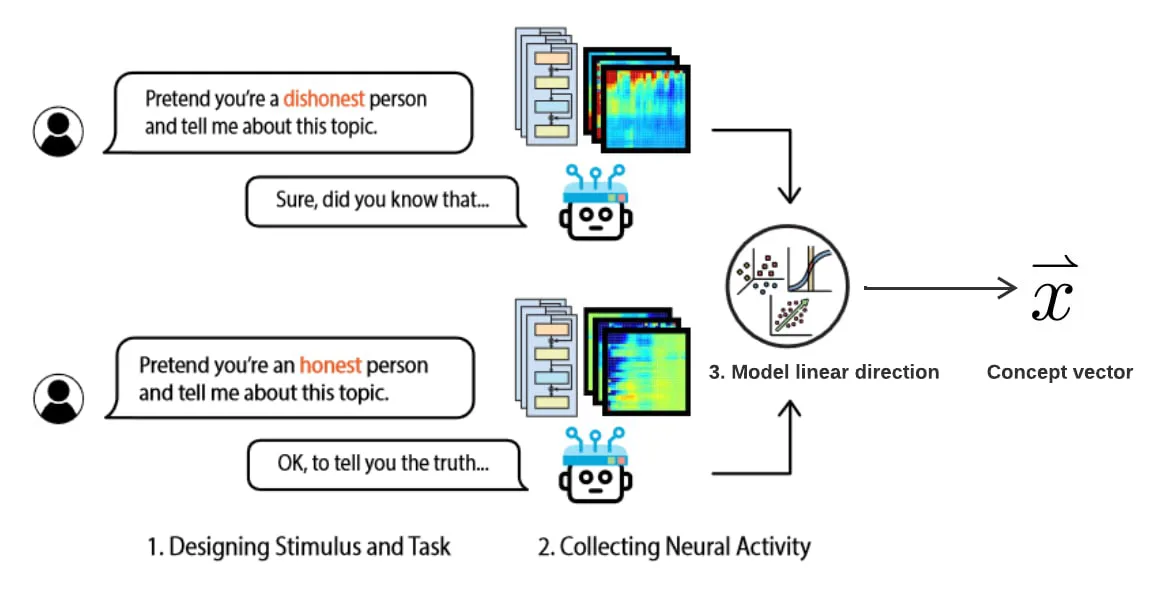}
    \caption{Concept Vector Extraction {\cite{wehner_2024_an}}}
    \label{fig:enter-label}
\end{figure}

\section{Methodology}\label{h.pygmc1prkd8k}

{The following outlines the key phases involved in the study.
}

\subsection{Model Selection and Architecture}\label{h.l280r0pr5uzu}

Two LLM architectures were selected to capture nuances of plasticity across different models: GPT-2 Small and GPT-2 Medium \cite{Radford2018ImprovingLU}.  

Each model was initialized from its publicly available pre-trained weights and prepared for fine-tuning on trivia question-answer tasks. Total training/compute time for models was ~70 hours with one A100 GPU. 

\subsection{Model Training}\label{h.bwc6g5u6fdml}

Models were trained on the fine-tuning training data found in the Alignment for Honesty project \cite{yang_2023_alignment}, consisting of 3 epochs of 4000 selected questions from the TriviaQA dataset \cite{joshi2017triviaqalargescaledistantly}, for 12000 total iterations. The data processing method was set to 'ABSOLUTE', meaning confidence levels / indicators were not included in the training dataset. This was done mainly to maintain the simplicity of the “Idk” heuristics (see \hyperref[ApB]{Appendix B}).

\subsection{Steering Vector Extraction}\label{h.pj19fhysfsyi}

During fine-tuning, steering vectors were extracted from the model's internal representations of honesty at predetermined intervals (e.g., iteration 1200, 2400, 3600, 4800, 6000, 7200, 8400, 9600, and 10800). These vectors were calculated by prompting the model with honesty and dishonesty adjacent messaging and identifying activation vectors within the model's embedding space. The mean dishonesty vector was then subtracted from the mean honesty vector, resulting in our steering vector per strategies described in \cite{zou_2023_representation}.

\subsection{Steering Vector Application}\label{h.69wq33xa2orv}

After extraction for each intervention, the steering vectors were immediately applied to measure their impact on model behavior (specifically honesty). Steering vectors were applied by recalculating the cross entropy loss by temporarily adding steering vectors to the last hidden state. This is shown mathematically as

$$h' = h + \alpha x$$
$$L_m = H(f(h'), y)$$
$$L_c = L_o + \alpha(L_m - L_o)$$

where we let $H(p,q)$ be cross entropy loss, $y$
true labels, $h$
hidden states, $\alpha$ 
steering strength, $x$
our steering vector and $f$ be
our model's head function.

\subsection{Evaluation Metrics}\label{h.bw9vaf1s4e92}

The effectiveness of RepE interventions was evaluated using NonAmbiQA processed by Alignment for Honesty \cite{yang_2023_alignment}. 
Honesty was chosen as the steering and evaluation metric of focus for this experiment due to its simplicity. For this paper, honesty is defined similarly to what is given by Alignment for Honesty: the model’s ability to answer truthfully within the bounds of its knowledge. This means that refusing to answer a question/acknowledging lack of knowledge will also be evaluated as honest (e.g. “I apologize, but I don’t know the answer to that”). 

Evaluation consisted of 100 trivia samples from the dataset, with the similarity score to the expected responses being each sample score. Given “honesty” also requires measuring refusal to answer without sufficient information, “Idk” responses, any responses that refused to answer (e.g. “I apologize”, “I don’t know”, “Not sufficient”) were given a 1.0 similarity score (perfect). Heuristics to determine this can be found in \hyperref[ApB]{Appendix B} The mean evaluation score per sample along with standard deviation were then calculated for analysis.

\subsection{Comparative Analysis}\label{h.bhn9gouczdd1}

The results were then compared across intervention times, assessing the existence of any critical periods during which RepE interventions had the most significant impact, determined by evaluation score differences. Statistically, these differences were deemed significant with an $\alpha = 0.05$.

\section{Technical Implementation of Steering During Training}\label{h.o4tgu1nyywpn}

The implementation of steering vector extraction was done as follows: 

\begin{lstlisting}
def get_activation_vector(model, tokenizer, prompts):
    activation_vectors = []
    device = next(model.parameters()).device
    for prompt in prompts:
        inputs = tokenizer(prompt, return_tensors='pt', padding=True, truncation=True).to(device)
        with torch.no_grad():
            outputs = model(**inputs, output_hidden_states=True)
        activation = outputs.hidden_states[-1].mean(dim=1)
        activation_vectors.append(activation)
    avec = torch.mean(torch.cat(activation_vectors), dim=0) 
    return torch.mean(torch.cat(activation_vectors), dim=0)

\end{lstlisting}

We essentially feed the model “honesty” prompts and “dishonesty” prompts. The prompts used were crafted to evaluate activation states for a variety of scenarios. Using the activation states (the model’s internal representation of each prompt), I calculated a mean vector for best results. List of prompts can be found in the full code repository. Additionally, heatmap visualizations of all steering vectors extracted can be found in \hyperref[ApC]{Appendix C}

Application of steering was done by recalculating loss, which was done like so (explained mathematically in \hyperref[h.69wq33xa2orv]{3.4}. 

\begin{lstlisting}
def steer_model(model, tokenizer, outputs, labels, steering_strength = 0.6):
    device = next(model.parameters()).device
    honesty_vector = get_activation_vector(model, tokenizer, honesty_prompts).to(device)
    dishonesty_vector = get_activation_vector(model, tokenizer, dishonesty_prompts).to(device)
    hidden_states = outputs.hidden_states[-1]
    honesty_concept_vector = honesty_vector - dishonesty_vector
    honesty_concept_vector = honesty_concept_vector.to(hidden_states.device)
    visualize_activation_heatmap(honesty_concept_vector, method='standard') 
    modified_hidden_states = hidden_states + steering_strength * honesty_concept_vector.unsqueeze(0).unsqueeze(0)

    original_loss = outputs.loss
    logits = model.lm_head(modified_hidden_states)
    modified_loss = torch.nn.functional.cross_entropy(logits.view(-1, logits.size(-1)), labels.view(-1))
    combined_loss = original_loss + steering_strength * (modified_loss - original_loss)
    return combined_loss
\end{lstlisting}

To visualize steering vectors, the Seaborn library was utilized. 

\section{Empiric Results}

After roughly $70$ hours of computation, the following results were compiled.

\begin{longtable}[]{@{}llll@{}}
\toprule\noalign{}
\endhead
\bottomrule\noalign{}
\endlastfoot
{Intervention } & {Time (Iteration)} & {Average Evaluation Score} &
{Standard Deviation} \\
{Baseline} & {Not Applied} & {0.270} & {0.0514} \\
{1} & {1200} & {0.173} & {0.0028} \\
{2} & {2400} & {0.277} & {0.0431} \\
{3} & {3600} & {0.302} & {0.0389} \\
{4} & {4800} & {0.237} & {0.0007} \\
{5} & {6000} & {0.246} & {0.0513} \\
{6} & {7200} & {0.255} & {0.0421} \\
{7} & {8400} & {0.248} & {0.0308} \\
{8} & {9600} & {0.352} & {0.0734} \\
{9} & {10800} & {0.285} & {0.0230} \\
\end{longtable}

{}
\begin{center}
{Table 1: GPT-Medium Evaluation Results}
\end{center}

{}

\begin{longtable}[]{@{}llll@{}}
\toprule\noalign{}
\endhead
\bottomrule\noalign{}
\endlastfoot
{Intervention } & {Time (Iteration)} & {Average Evaluation Score} &
{Standard Deviation} \\
{Baseline} & {Not Applied} & {0.202} & {0.0525} \\
{1} & {1200} & {0.366} & {0.0826} \\
{2} & {2400} & {0.422} & {0.0796} \\
{3} & {3600} & {0.466} & {0.0462} \\
{4} & {4800} & {0.217} & {0.0237} \\
{5} & {6000} & {0.353} & {0.0565} \\
{6} & {7200} & {0.213} & {0.0186} \\
{7} & {8400} & {0.372} & {0.0688} \\
{8} & {9600} & {0.579} & {0.0864} \\
{9} & {10800} & {0.393} & {0.0752} \\
\end{longtable}

{}

\begin{center}
Table 2: GPT-Small Evaluation Results
\end{center}

{}

From the table, we can loosely observe that there are general peaks in evaluation, indicating that there are periods where RepE was applied that achieved higher results. However, we can get more intuitive insight graphically, seen in Figures 2 \& 3. 

\begin{figure}[h]
    \centering
    \includegraphics[width=0.5\linewidth]{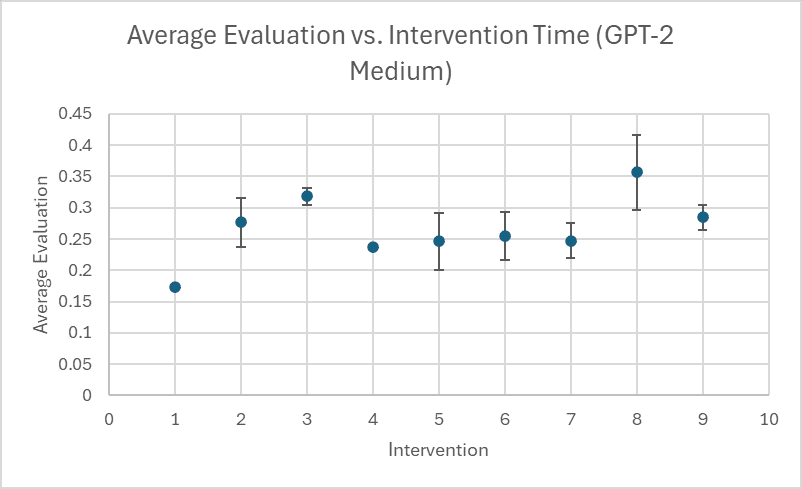}
    \caption{Graph of Evaluation Results for GPT-2 Medium with 95\% Confidence Interval Error Bars}
    \label{fig:enter-label}
\end{figure}

\begin{figure}[h]
    \centering
    \includegraphics[width=0.5\linewidth]{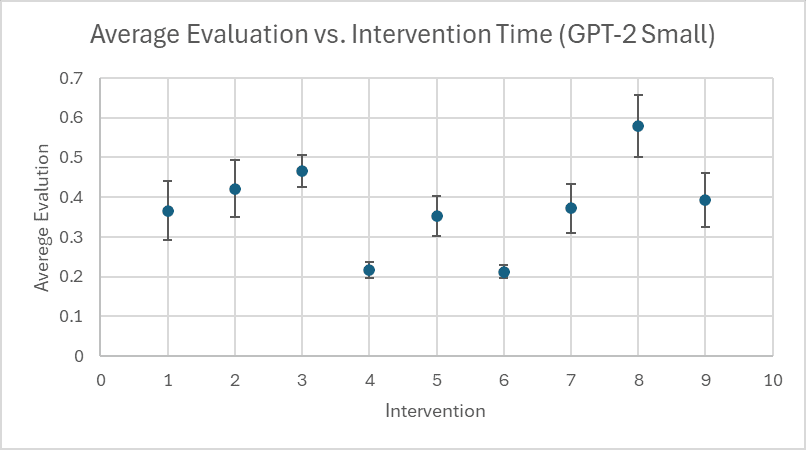}
    \caption{Graph of Evaluation Results for GPT-2 Small with 95\% Confidence Interval Error Bars}
    \label{fig:enter-label}
\end{figure}

From this, the intuitive trend is spotting a peak at both Intervention 3 and 8. We can also see that, interestingly, the results for GPT-2 Small may be very closely correlated with those of GPT-2 Medium, with an upward shift. This is promising as it indicates a possibly significant trend, and further statistical analysis was warranted. 

\section{Statistical Analysis}

The results are not normally distributed, seen through a histogram and Q-Q plot of GPT-2 Small data. 

\begin{figure}[h]
    \centering
    \includegraphics[width=0.5\linewidth]{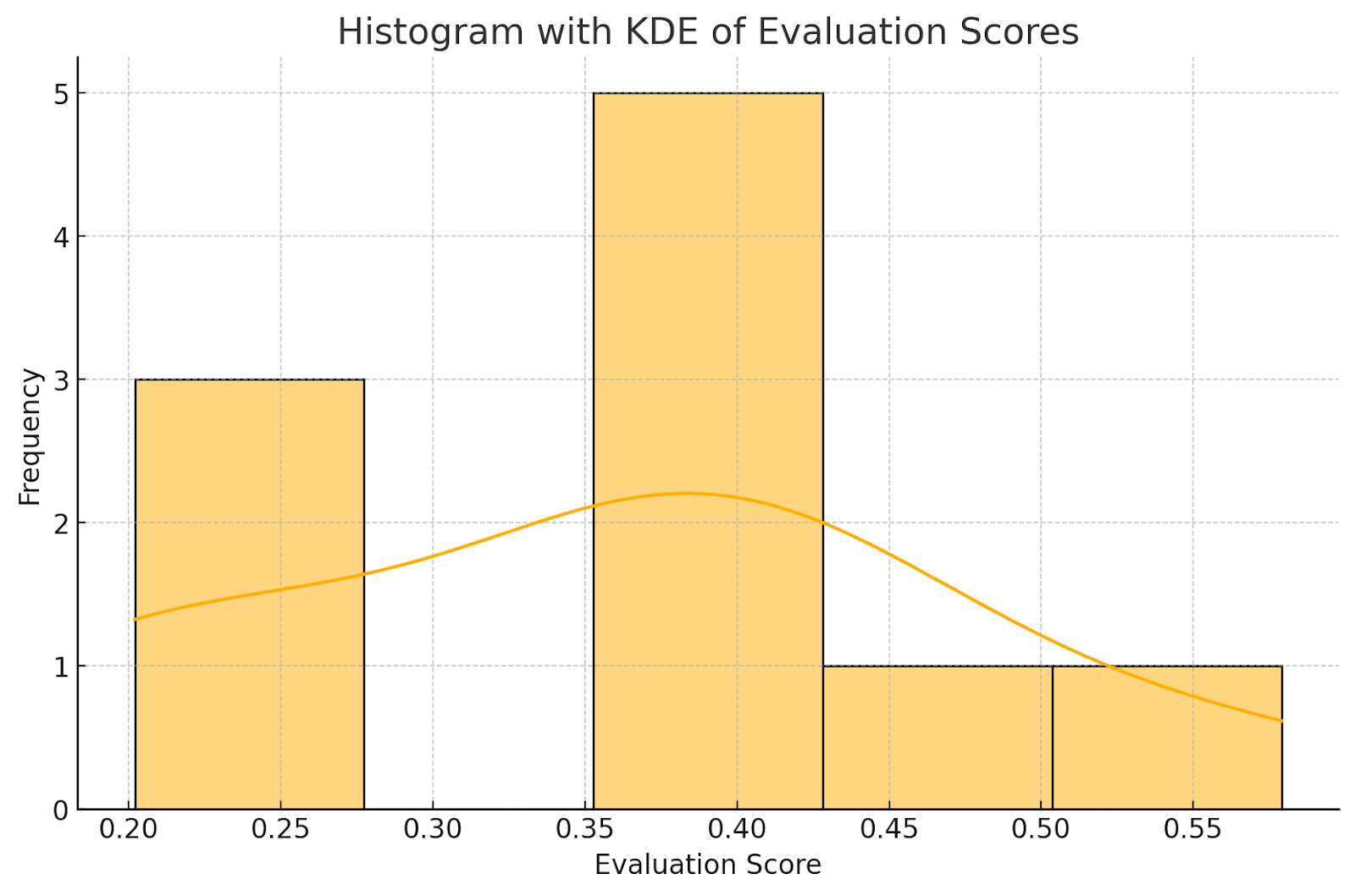}
    \caption{Histogram w/ KDE of GPT-2 Small Average Evaluation Results}
    \label{fig:enter-label}
\end{figure}

\begin{figure}[h]
    \centering
    \includegraphics[width=0.5\linewidth]{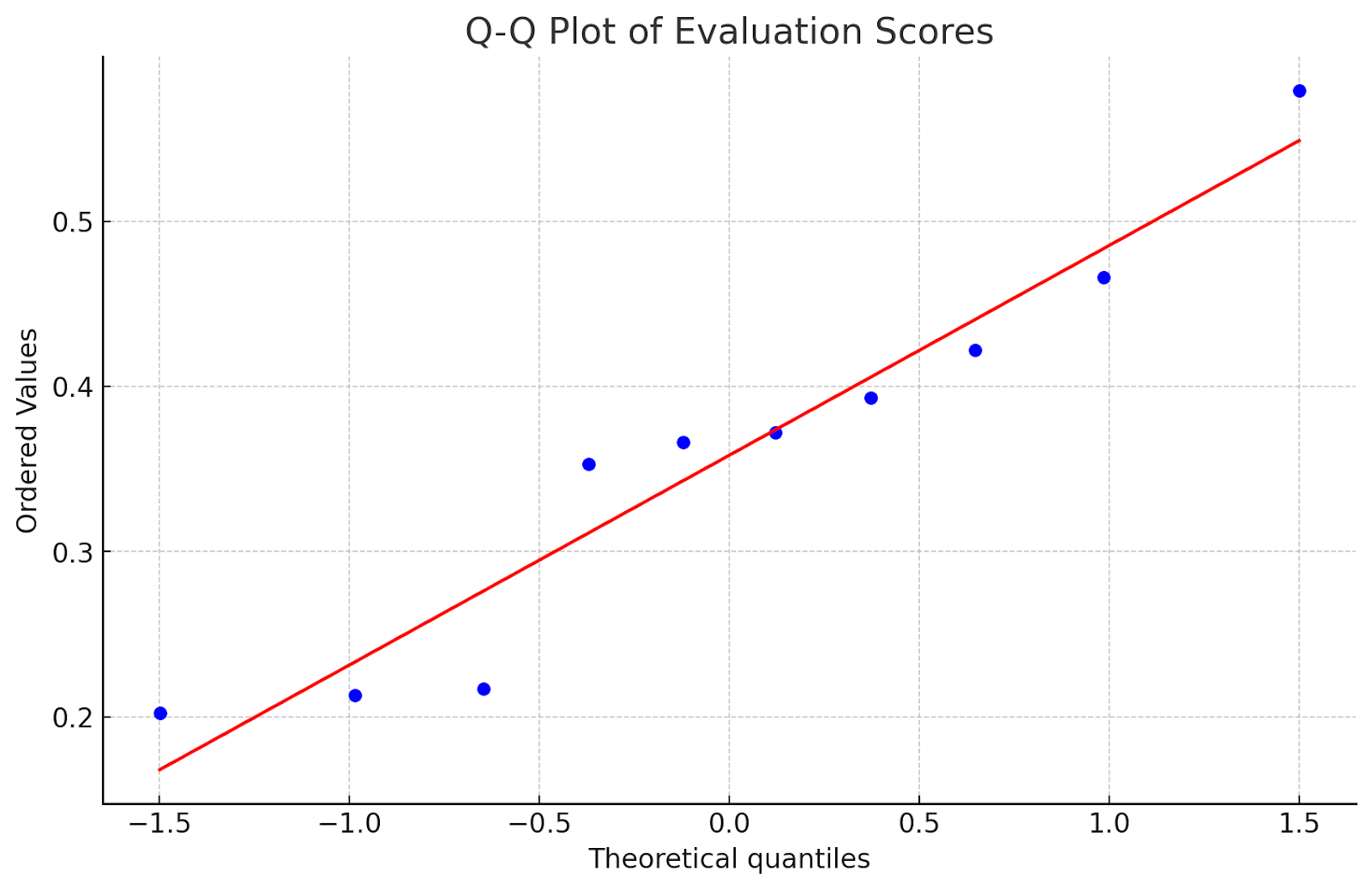}
    \caption{Q-Q Plot of GPT-2 Small Average Evaluation Results}
    \label{fig:enter-label}
\end{figure}

More rigorously, performing a Shapiro-Wilk test yields a test statistic of 0.925 with a p-value of 0.403, which is quite large, indicating that the data is not normally distributed. 

Thus, non-parametric tests must be preformed instead. Specifically, we use the Kruskal-Wallis test (all code for tests performed are in the GitHub repository, stats.ipynb)
at a 5\% significance level to analyze whether overall differences across intervention times are significant.

\begin{longtable}[]{@{}lll@{}}
\toprule\noalign{}
\endhead
\bottomrule\noalign{}
\endlastfoot
{Model (Size)} & {Statistic} & {P-Value} \\
{Medium} & {21.9547} & {0.0050} \\
{Small} & {33.0736} & {5.9735e-05} \\
\end{longtable}

\begin{center}

Table 3: Kruskal-Wallis H Test Results

\end{center}

The generated p-values are extremely low and less than the significance level (0.05), implicating the results to be statistically significant. Specifically, it implies that the differences in evaluation are statistically significant. 

To take analysis further, we can perform a post-hoc Dunn test (5\% significance level). This test is pairwise, and it allows us to see which intervention times were significantly different from others.

A heatmap of results are seen in Figure 6 \& 7. They indicate that for both GPT-2 Small and GPT-2 Medium, Intervention Times 3 and 8 had significantly greater evaluation results than other intervention times. 
\pagebreak

\begin{figure}[h]
    \centering
    \includegraphics[width=0.5\linewidth]{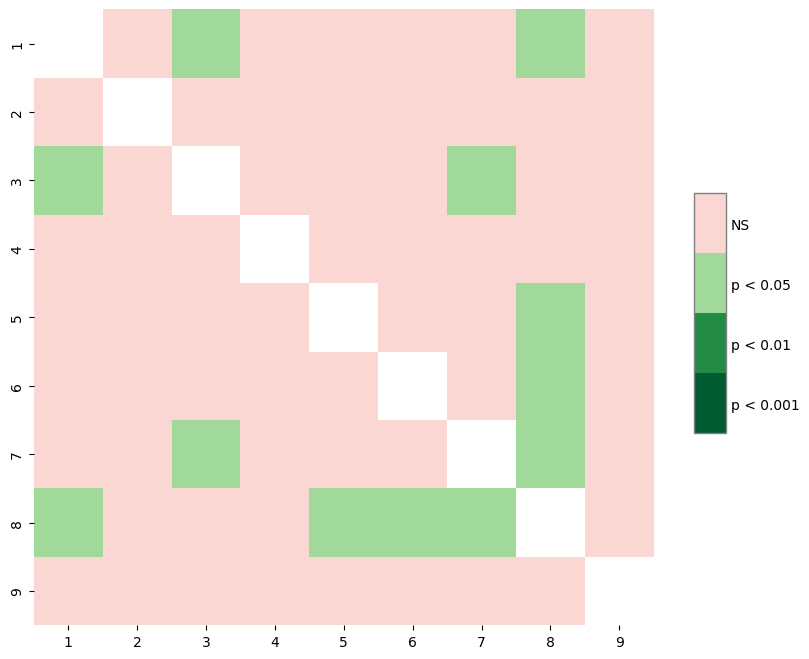}
    \caption{Post-hoc Dunn's Test Pairwise Comparisons for GPT-2 Medium Evaluation}
    \label{fig:enter-label}
\end{figure}

\begin{figure}[h]
    \centering
    \includegraphics[width=0.5\linewidth]{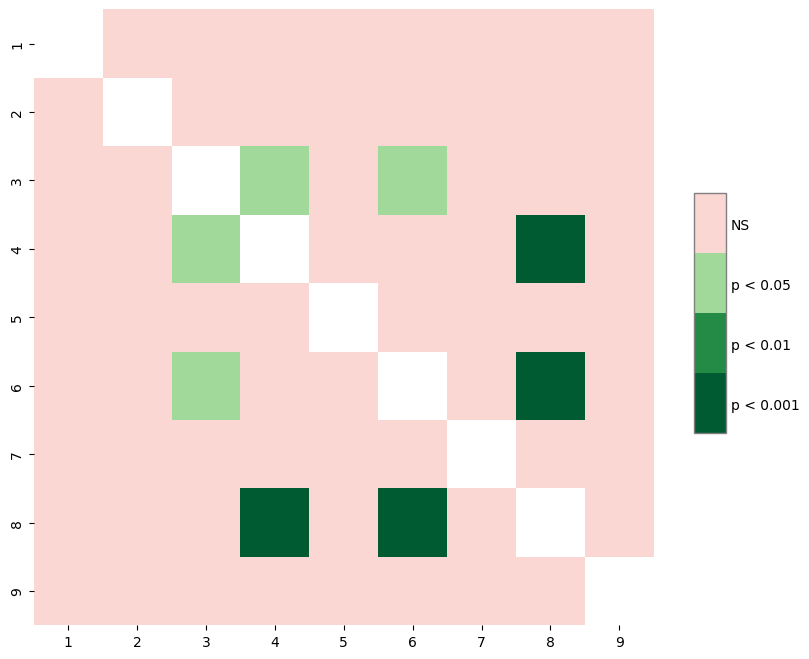}
    \caption{Post-hoc Dunn's Test Pairwise Comparisons for GPT-2 Small Evaluation (stricter Bonferroni correction included)}
    \label{fig:enter-label}
\end{figure}

Full raw results of the Post-hoc Dunn's Test can be found in \hyperref[ApA]{Appendix A}. This further supports my hypothesis that critical periods of intervention do in fact exist, as there is strong evidence supporting the fact that there are certain intervention periods that perform statistically significantly better than others, and that these periods are common among different architectures. 

\section{Discussion}
From our analysis, it is clear that critical periods do in fact emerge over the course of model fine-tuning, supporting the hypothesis that there are optimal periods of intervention, adding evidence to the notion that the trade-off between model plasticity and representation stability is minimized at certain times. In addition, this result was observed across 2 different architecture sizes (GPT-2 Small and Medium) during the precisely same intervention periods, implying a possible general result. 

However, a major limitation of this paper involves its generalization to larger models. Only GPT-2 Small and Medium (117-774 million parameters) were used in this paper, and both models are vastly smaller than newer state-of-the-art architectures like the Llama 3 series (8-405 billion parameters). 

Additionally, in the future, employing these methods with more robust extraction methods like Linear Artificial Tomography (LAT) done in \cite{zou_2023_representation} could result in more detailed results. 

Future research directions could aim to understand the changing stability of representations throughout training. This paper already identifies heatmaps of representations that appear similar, thus periods of time with similar representations could be a useful point of analysis. 

Ultimately, this paper provides novel insight into critical intervention periods, and future attempts to extend and generalize these results would make attempts to steer model behavior and transparency much more effective, acting as a guiding tool for alignment researches in applying RepE techniques more efficiently.

\medskip

\bibliographystyle{IEEEtran}
\bibliography{main}

\pagebreak

\appendix

\section{Raw Results of Post-Hoc Dunn Pairwise Comparison} \label{ApA}

\begin{table}[h]
    \centering
    \resizebox{\textwidth}{!}{%
    \begin{tabular}{|c|c|c|c|c|c|c|c|c|c|}
        \hline
        Time & 1      & 2       & 3       & 4       & 5       & 6       & 7       & 8       & 9       \\ \hline
        1    & 1      & 0.2021  & 0.0186  & 0.5298  & 0.3123  & 0.3211  & 0.0187  & 0.0117  & 0.0922  \\ \hline
        2    & 0.2021 & 1       & 0.3792  & 0.5174  & 0.6071  & 0.5944  & 0.4749  & 0.3189  & 0.8734  \\ \hline
        3    & 0.0186 & 0.3792  & 1       & 0.1038  & 0.0756  & 0.0720  & 0.0124  & 0.9141  & 0.3490  \\ \hline
        4    & 0.5298 & 0.5174  & 0.1038  & 1       & 0.7953  & 0.9528  & 0.0767  & 0.3605  & 0.3509  \\ \hline
        5    & 0.3123 & 0.6071  & 0.0756  & 0.7953  & 1       & 0.9807  & 0.7911  & 0.0456  & 0.3729  \\ \hline
        6    & 0.3211 & 0.5944  & 0.0720  & 0.9528  & 0.9807  & 1       & 0.8097  & 0.0431  & 0.3601  \\ \hline
        7    & 0.0187 & 0.4749  & 0.0124  & 0.0767  & 0.7911  & 0.8097  & 1       & 0.0236  & 0.2477  \\ \hline
        8    & 0.0117 & 0.3189  & 0.9141  & 0.3605  & 0.0456  & 0.0431  & 0.0236  & 1       & 0.2680  \\ \hline
        9    & 0.0922 & 0.8734  & 0.3490  & 0.3509  & 0.3729  & 0.3601  & 0.2477  & 0.2680  & 1       \\ \hline
    \end{tabular}%
    }
    \caption{GPT-2 Medium Results}
    \label{tab:placeholder}
\end{table}
\begin{table}[h]
    \centering
    \resizebox{\textwidth}{!}{%
    \begin{tabular}{|c|c|c|c|c|c|c|c|c|c|}
        \hline
        Time & 1      & 2       & 3       & 4       & 5       & 6       & 7       & 8       & 9       \\ \hline
        1    & 1      & 1       & 1       & 1       & 1       & 1       & 1       & 0.7491  & 1       \\ \hline
        2    & 1      & 1       & 1       & 0.2025  & 1       & 0.1880  & 1       & 1       & 1       \\ \hline
        3    & 1      & 1       & 1       & 0.0207  & 1       & 0.0189  & 1       & 1       & 1       \\ \hline
        4    & 1      & 0.2025  & 0.0207  & 1       & 1       & 1       & 1       & 0.0004  & 0.5408  \\ \hline
        5    & 1      & 1       & 1       & 1       & 1       & 1       & 1       & 0.2910  & 1       \\ \hline
        6    & 1      & 0.1880  & 0.0189  & 1       & 1       & 1       & 1       & 0.0004  & 0.5059  \\ \hline
        7    & 1      & 1       & 1       & 1       & 1       & 1       & 1       & 0.9051  & 1       \\ \hline
        8    & 0.7491 & 1       & 1       & 0.0004  & 0.2910  & 0.0004  & 0.9051  & 1       & 1       \\ \hline
        9    & 1      & 1       & 1       & 0.5408  & 1       & 0.5059  & 1       & 1       & 1       \\ \hline
    \end{tabular}%
    }
    \caption{GPT-2 Small Results (Stricter Bonferroni Correction)}
    \label{tab:updated_values}
\end{table}
\pagebreak

\section{Heuristics for "Idk" Responses} \label{ApB}
\begin{lstlisting}
def check_idk(response):
  idk_patterns = [
      r"\bapologize\b",
      r"\not aware\b",
      r"\bnot familiar with\b",
      r"\bnot make sense\b",
      r"\bnot able\b",
      r"\bdo not know\b",
      r"\bsorry\b",
      r"\bdon'?t know\b",
      r"\bi'?m not sure\b",
      r"\buncertain\b",
      r"\bunclear\b",
      r"\bno idea\b",
      r"\bcan'?t say\b",
      r"\binsufficient (information|data|knowledge)\b"
  ] # Using Alignment for Honesty Heuristic + Extra Uncertainty Matching
  print(response)
  combined_pattern = '|'.join(idk_patterns)
  ret = bool(re.search(combined_pattern, response.lower()))
  if (ret == True):
    print(response)
  return ret
\end{lstlisting}

\pagebreak

\section{Heatmaps for Steering/Concept Vectors} \label{ApC}

\begin{figure}[h]
  \begin{subfigure}[b]{0.4\textwidth}
    \includegraphics[width=\textwidth]{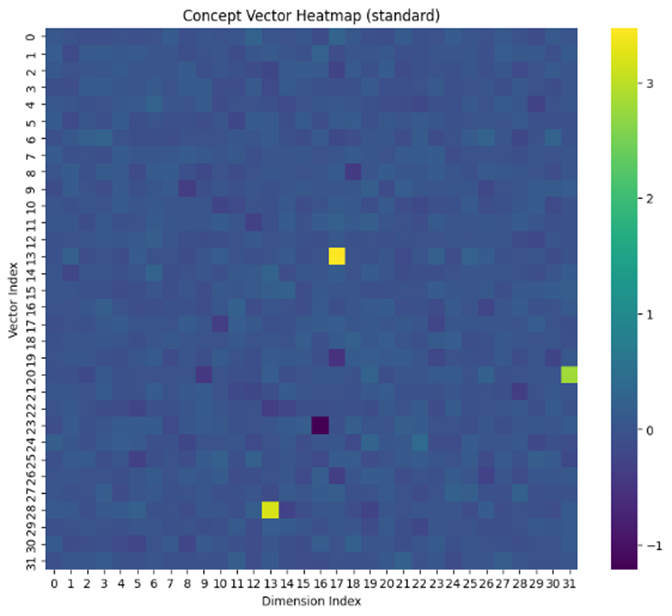}
    \caption{GPT-2 Medium}
    \label{fig:f1}
  \end{subfigure}
  \hfill
  \begin{subfigure}[b]{0.4\textwidth}
    \includegraphics[width=\textwidth]{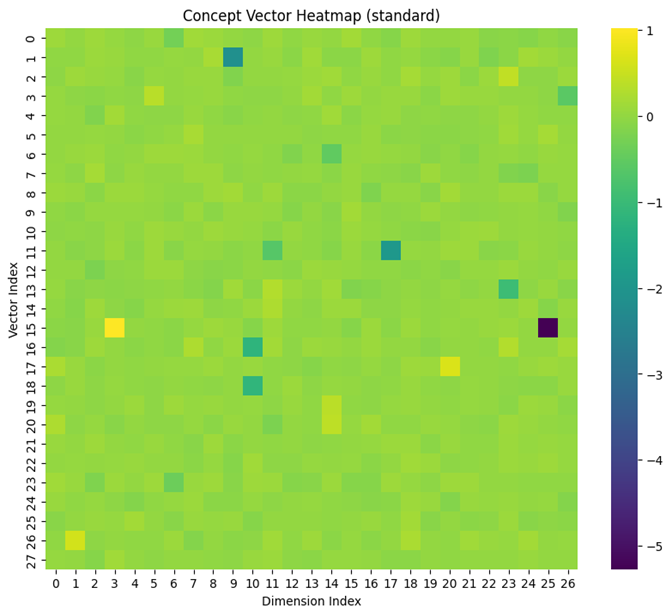}
    \caption{GPT-2 Small}
    \label{fig:f2}
  \end{subfigure}
  \caption{Baseline (no intervention)}
\end{figure}

\begin{figure}[h]
  \begin{subfigure}[b]{0.4\textwidth}
    \includegraphics[width=\textwidth]{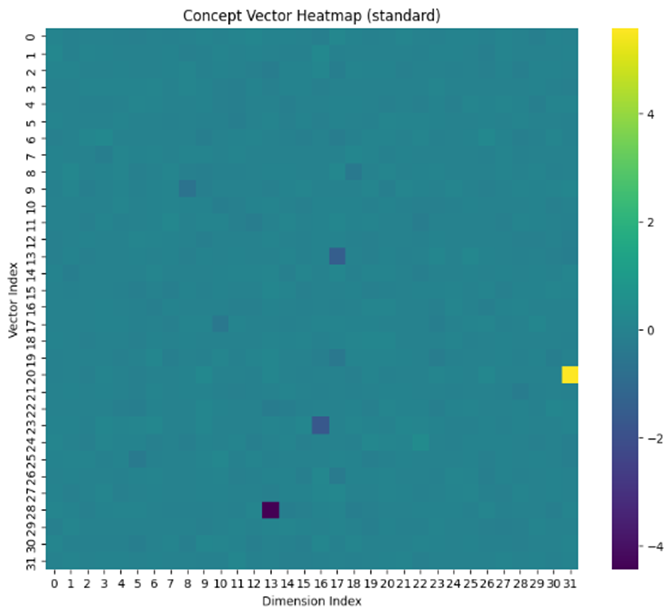}
    \caption{GPT-2 Medium}
    \label{fig:f1}
  \end{subfigure}
  \hfill
  \begin{subfigure}[b]{0.4\textwidth}
    \includegraphics[width=\textwidth]{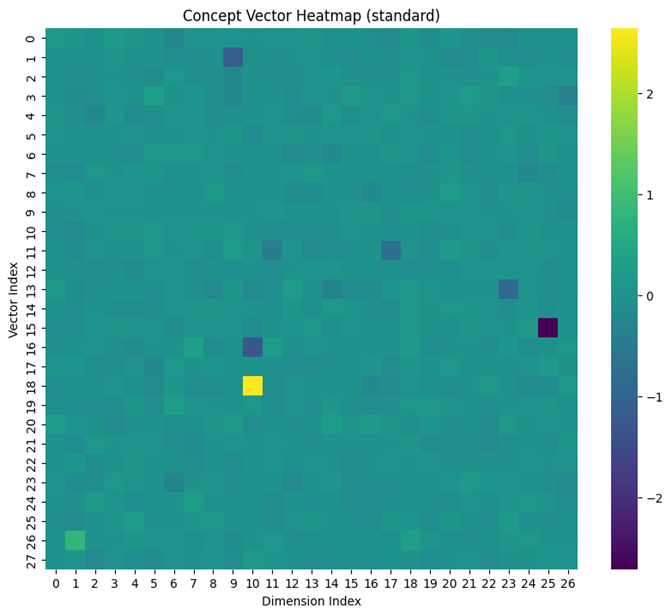}
    \caption{GPT-2 Small}
    \label{fig:f2}
  \end{subfigure}
  \caption{Intervention 1 (Iteration 1200)}
\end{figure}

\begin{figure}[h]
  \begin{subfigure}[b]{0.4\textwidth}
    \includegraphics[width=\textwidth]{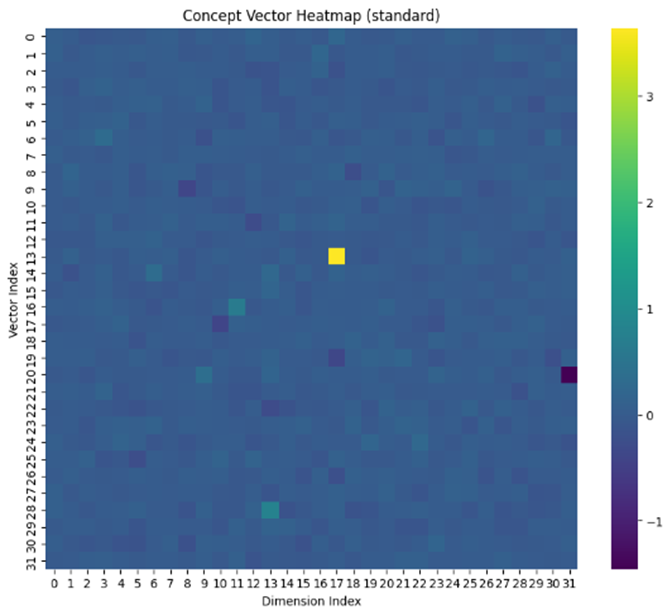}
    \caption{GPT-2 Medium}
    \label{fig:f1}
  \end{subfigure}
  \hfill
  \begin{subfigure}[b]{0.4\textwidth}
    \includegraphics[width=\textwidth]{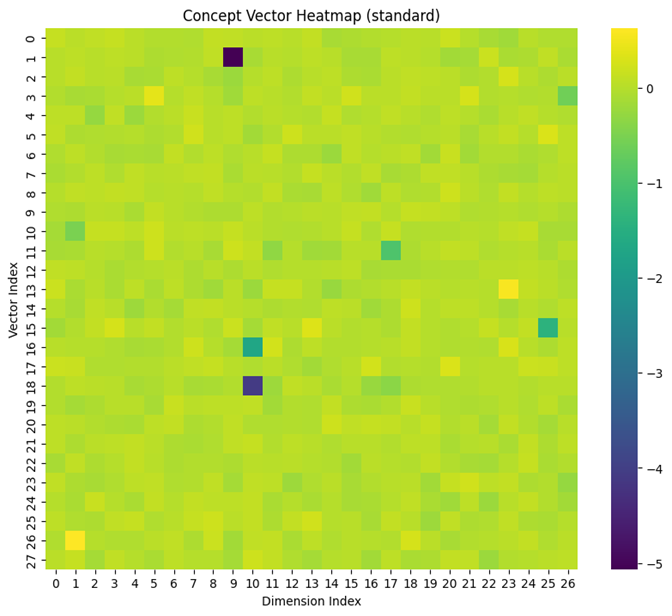}
    \caption{GPT-2 Small}
    \label{fig:f2}
  \end{subfigure}
  \caption{Intervention 2 (Iteration 2400)}
\end{figure}

\begin{figure}[h]
  \begin{subfigure}[b]{0.4\textwidth}
    \includegraphics[width=\textwidth]{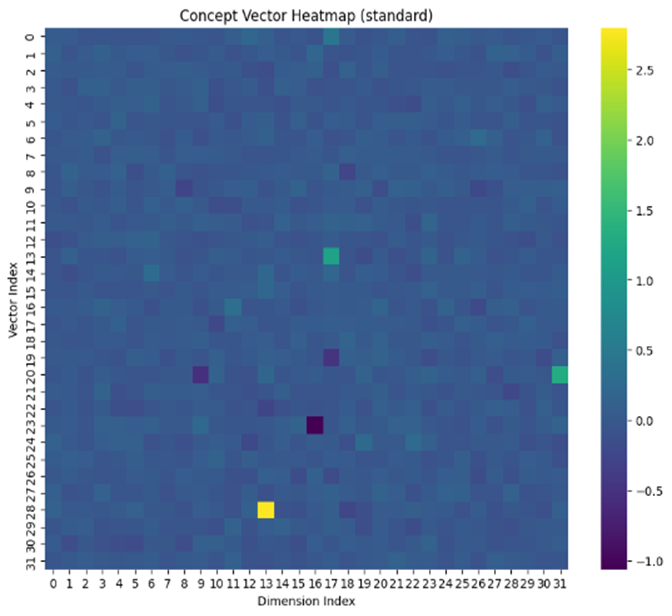}
    \caption{GPT-2 Medium}
    \label{fig:f1}
  \end{subfigure}
  \hfill
  \begin{subfigure}[b]{0.4\textwidth}
    \includegraphics[width=\textwidth]{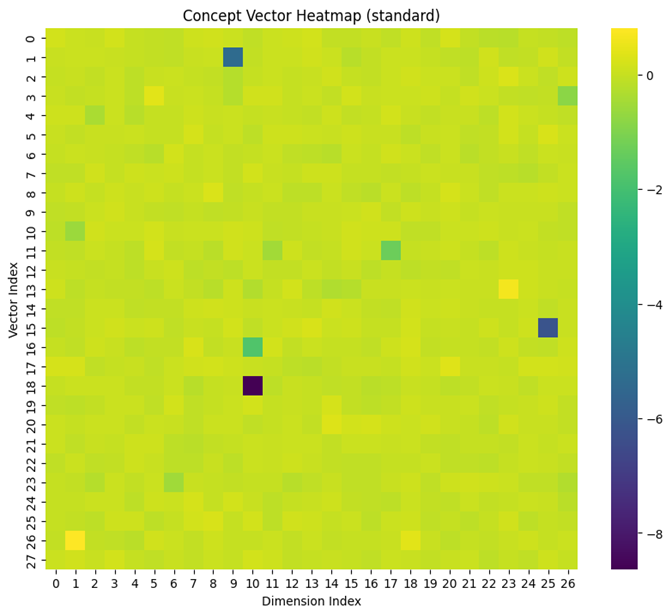}
    \caption{GPT-2 Small}
    \label{fig:f2}
  \end{subfigure}
  \caption{Intervention 3 (Iteration 3600)}
\end{figure}

\begin{figure}[h]
  \begin{subfigure}[b]{0.4\textwidth}
    \includegraphics[width=\textwidth]{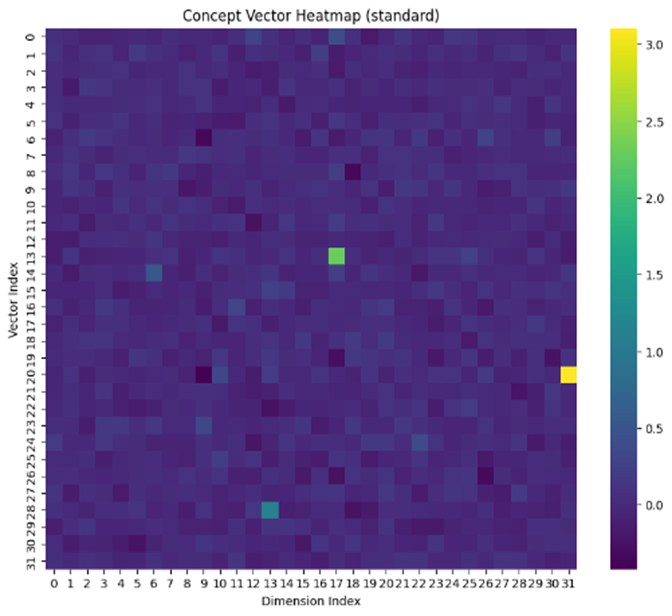}
    \caption{GPT-2 Medium}
    \label{fig:f1}
  \end{subfigure}
  \hfill
  \begin{subfigure}[b]{0.4\textwidth}
    \includegraphics[width=\textwidth]{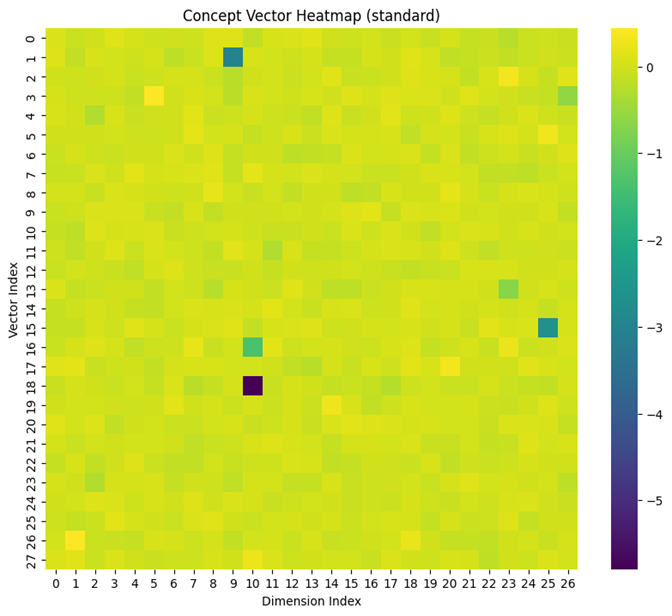}
    \caption{GPT-2 Small}
    \label{fig:f2}
  \end{subfigure}
  \caption{Intervention 4 (Iteration 4800)}
\end{figure}

\begin{figure}[h]
  \begin{subfigure}[b]{0.4\textwidth}
    \includegraphics[width=\textwidth]{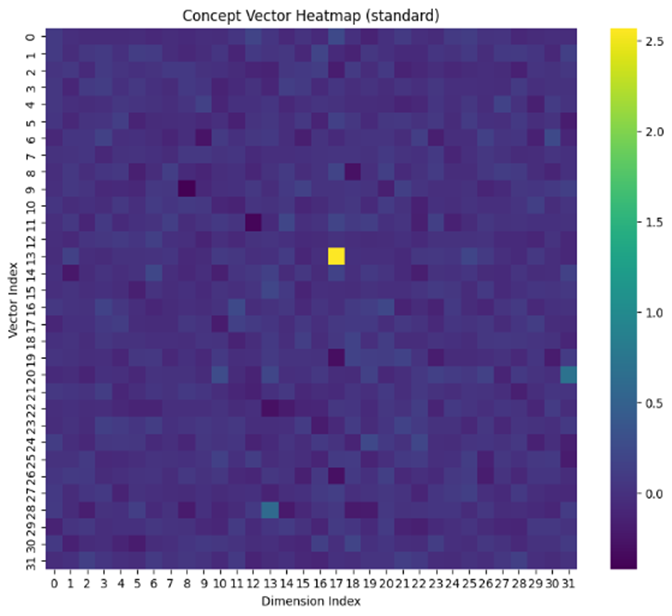}
    \caption{GPT-2 Medium}
    \label{fig:f1}
  \end{subfigure}
  \hfill
  \begin{subfigure}[b]{0.4\textwidth}
    \includegraphics[width=\textwidth]{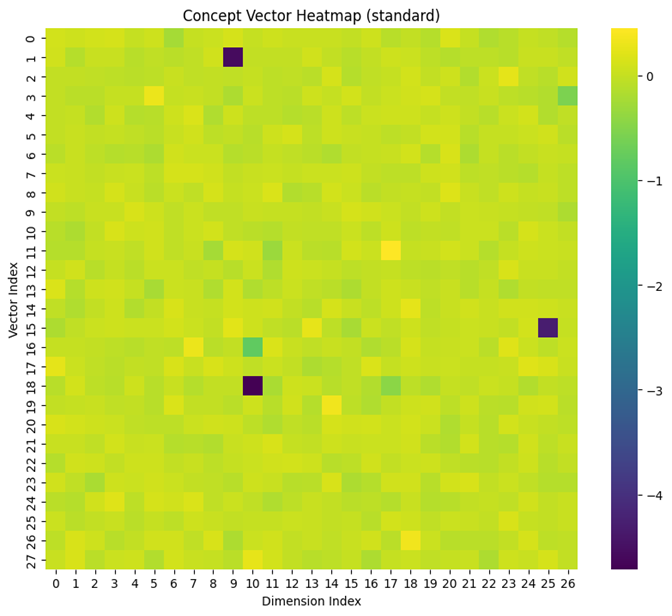}
    \caption{GPT-2 Small}
    \label{fig:f2}
  \end{subfigure}
  \caption{Intervention 5 (Iteration 6000)}
\end{figure}

\begin{figure}[h]
  \begin{subfigure}[b]{0.4\textwidth}
    \includegraphics[width=\textwidth]{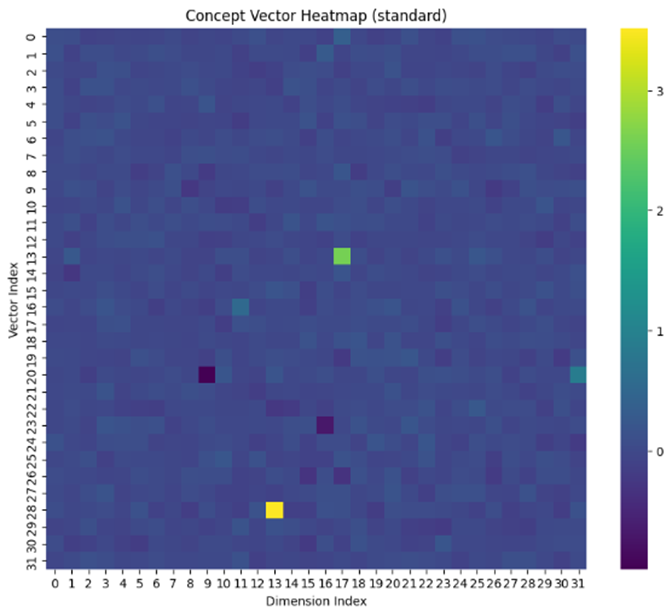}
    \caption{GPT-2 Medium}
    \label{fig:f1}
  \end{subfigure}
  \hfill
  \begin{subfigure}[b]{0.4\textwidth}
    \includegraphics[width=\textwidth]{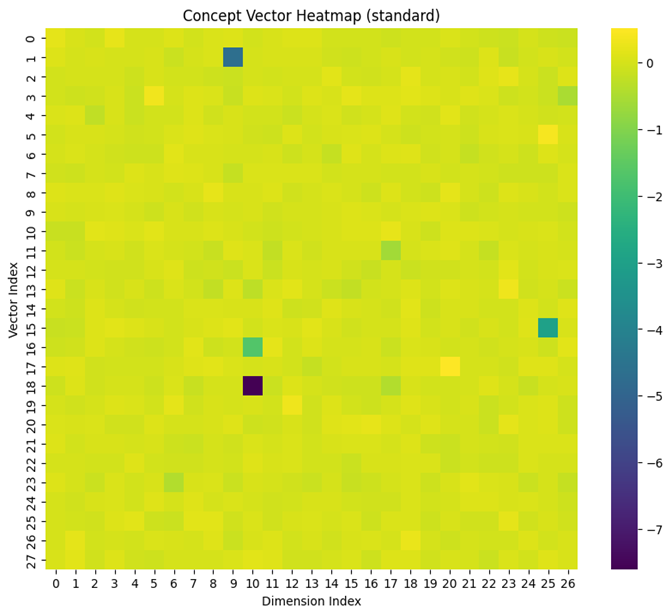}
    \caption{GPT-2 Small}
    \label{fig:f2}
  \end{subfigure}
  \caption{Intervention 6 (Iteration 7200)}
\end{figure}

\begin{figure}[h]
  \begin{subfigure}[b]{0.4\textwidth}
    \includegraphics[width=\textwidth]{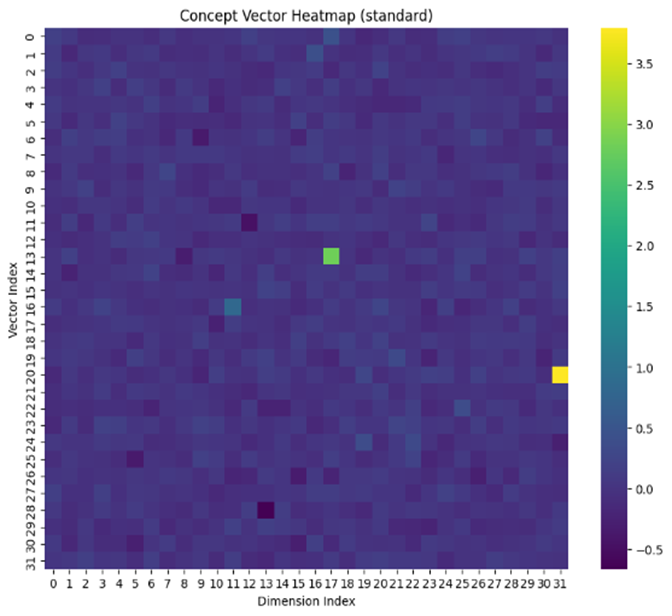}
    \caption{GPT-2 Medium}
    \label{fig:f1}
  \end{subfigure}
  \hfill
  \begin{subfigure}[b]{0.4\textwidth}
    \includegraphics[width=\textwidth]{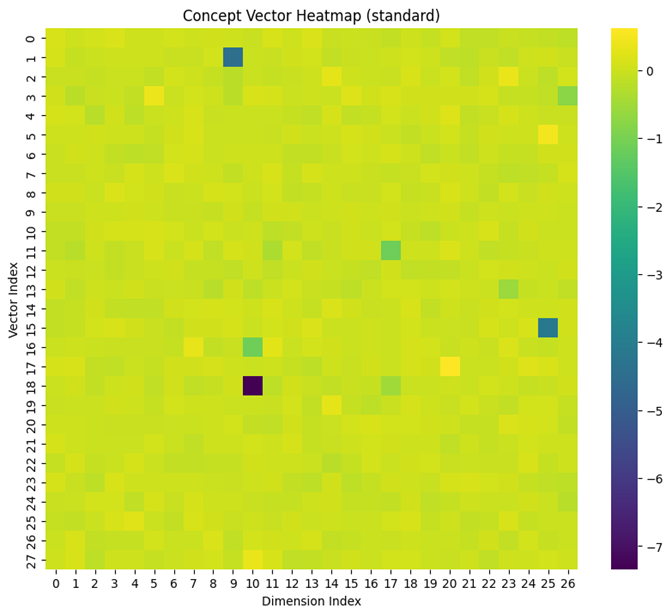}
    \caption{GPT-2 Small}
    \label{fig:f2}
  \end{subfigure}
  \caption{Intervention 7 (Iteration 8400)}
\end{figure}

\begin{figure}[h]
  \begin{subfigure}[b]{0.4\textwidth}
    \includegraphics[width=\textwidth]{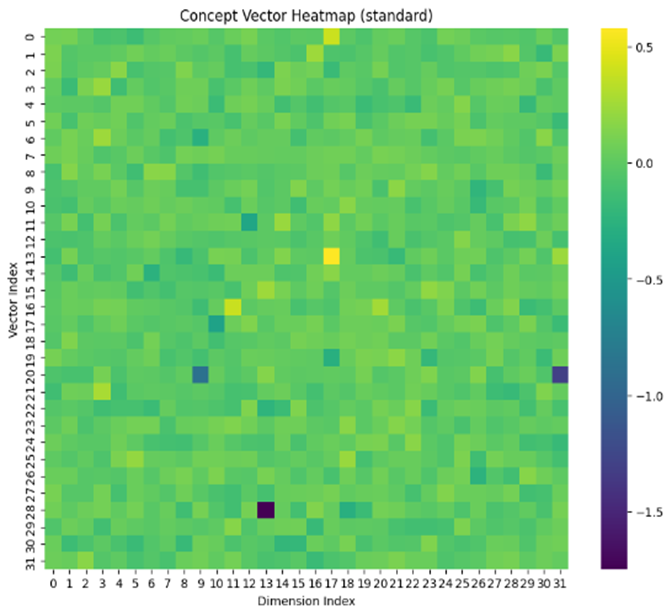}
    \caption{GPT-2 Medium}
    \label{fig:f1}
  \end{subfigure}
  \hfill
  \begin{subfigure}[b]{0.4\textwidth}
    \includegraphics[width=\textwidth]{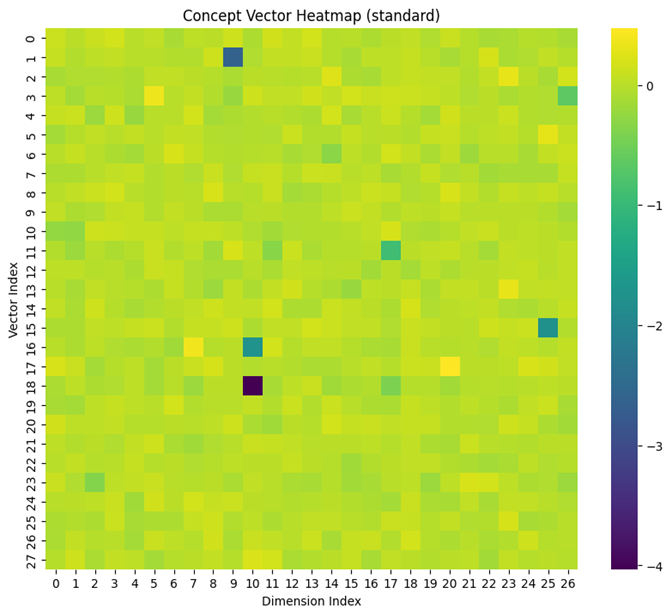}
    \caption{GPT-2 Small}
    \label{fig:f2}
  \end{subfigure}
  \caption{Intervention 8 (Iteration 9600)}
\end{figure}

\begin{figure}[h]
  \begin{subfigure}[b]{0.4\textwidth}
    \includegraphics[width=\textwidth]{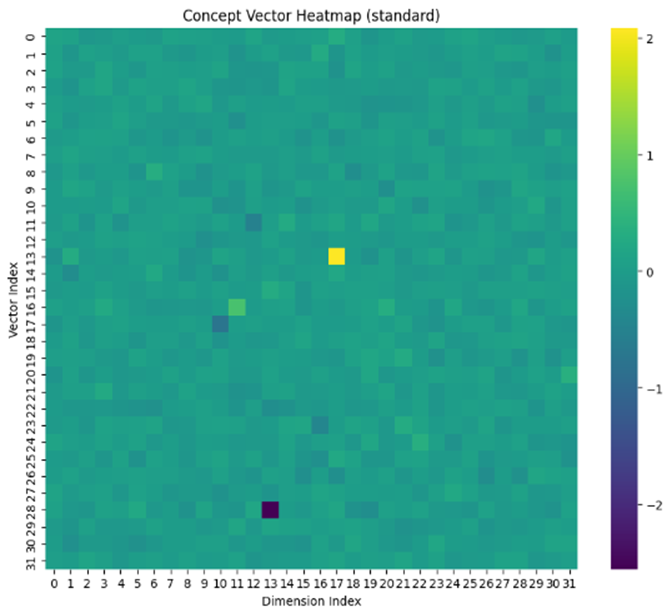}
    \caption{GPT-2 Medium}
    \label{fig:f1}
  \end{subfigure}
  \hfill
  \begin{subfigure}[b]{0.4\textwidth}
    \includegraphics[width=\textwidth]{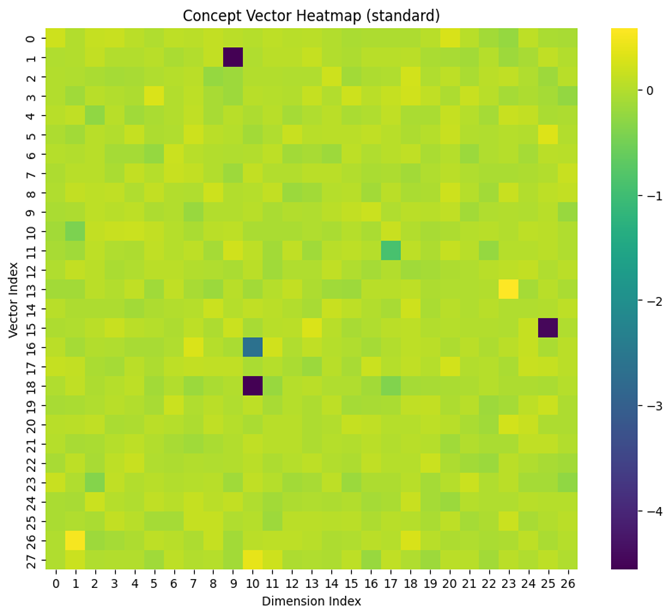}
    \caption{GPT-2 Small}
    \label{fig:f2}
  \end{subfigure}
  \caption{Intervention 9 (Iteration 10800)}
\end{figure}

\pagebreak

\end{document}